\documentclass[final]{cvpr}

\usepackage{times}
\usepackage{epsfig}
\usepackage{graphicx}
\usepackage{amsmath}
\usepackage{amssymb}

\usepackage{color}
\usepackage{graphicx}
\usepackage{comment}
\usepackage{adjustbox}
\usepackage{color}
\usepackage{epsfig}
\usepackage{graphicx}
\usepackage{booktabs}
\usepackage{colortbl}
\usepackage[numbers,sort,compress]{natbib}
\usepackage{url}
\usepackage{caption}
\usepackage{subcaption}
\usepackage{pifont}% http://ctan.org/pkg/pifont

\def\mL{\mathcal{L}}

\def\1n{\mathbf{1}_n}
\def\0{\mathbf{0}}
\def\1{\mathbf{1}}

\newcommand{\denselist}{\itemsep -1pt}

\definecolor{pink}{rgb}{0.9,0.5,0.5}
\definecolor{purple}{rgb}{0.5, 0.4, 0.8}   
\definecolor{gray}{rgb}{0.3, 0.3, 0.3}
\definecolor{mygreen}{rgb}{0.2, 0.6, 0.2}

\definecolor{greena}{rgb}{0.4, 0.5, 0.1}

\definecolor{bluea}{rgb}{0, 0.4, 0.6}

\definecolor{reda}{rgb}{0.6, 0.2, 0.1}

\newcommand{\cm}[1]{}

\newcommand{\mhoai}[1]{{\color{magenta}\textbf{[MH: #1]}}}

\newcommand{\myheading}[1]{\vspace{1ex}\noindent \textbf{#1}}

\newif\ifshowsolution
\showsolutiontrue

\ifshowsolution

\else

\fi

\newcommand{\vr}[1]{\textcolor{blue}{[VR: {#1}]}}

\newcommand{\Eref}[1]{Eq.~(\ref{#1})}
\newcommand{\Fref}[1]{Fig.~\ref{#1}}
\newcommand{\Tref}[1]{Table~\ref{#1}}

\newcommand{\cmark}{\ding{51}}%
\newcommand{\xmark}{\ding{55}}%
\usepackage[pagebackref=true,breaklinks=true,colorlinks,bookmarks=false]{hyperref}

\def\cvprPaperID{3902} % *** Enter the CVPR Paper ID here
\def\confYear{CVPR 2021}

\begin{document}

%%%%%%%%% TITLE
\title{Learning To Count Everything}

% \and

\author{Viresh Ranjan$^{1}$ \quad Udbhav Sharma$^{1}$  \quad Thu Nguyen$^{2}$ \quad Minh Hoai$^{1,2}$ \\
$^1$Stony Brook University, USA\\
$^2$VinAI Research, Hanoi, Vietnam 
}

\maketitle

%%%%%%%%% ABSTRACT

\begin{abstract}
Existing works on visual counting primarily focus on one specific category at a time, such as people, animals, and cells. In this paper, we are interested in counting everything, that is to count objects from any category given only a few annotated instances from that category.
To this end, we pose counting as a few-shot regression task. 
To tackle this task, we present a novel method that takes a query image together with a few exemplar objects from the query image and predicts a density map for the presence of all objects of interest in the query image. We also present a novel adaptation strategy to adapt our network to any novel visual category at test time, using only a few exemplar objects from the novel category.  We also introduce a dataset of 147 object categories containing over 6000 images that are suitable for the few-shot counting task. The images are annotated with two types of annotation, dots and bounding boxes, and they can be used for developing few-shot counting models. Experiments on this dataset shows that our method outperforms several state-of-the-art object detectors and few-shot counting approaches. Our code and dataset can be found at \url{https://github.com/cvlab-stonybrook/LearningToCountEverything}. 
%, and show that our model outperforms all of the existing approaches.

% To this end, we pose counting as a few-shot regression task and develop a novel method for this task. 

% We collect a dataset of 147 object categories containing over 6000 images that are suitable for the few-shot counting task. The images are annotated with two types of annotation, dots and bounding boxes, and they can be used for developing few-shot visual counting models. 

% To tackle the task of few-shot counting, we capitalize on the idea of self-similarity and present an architecture that takes a query image together with a few exemplar objects from the query image, and predicts a density map for the presence of all objects of interest in the query image. We also present a novel adaptation strategy to adapt our network to any novel visual category at test time, using only a few exemplar objects from the novel category.  We compare our approach with several state-of-the-art object detectors and few shot counting approaches, and show that our model outperforms all of the existing approaches. We also show a surprising result that our proposed approach performs reasonably well on CARPK car counting dataset without using any training data specific to the car counting task. 
\end{abstract}

\section{Introduction}
Humans can count objects from most of the visual object categories with ease, while  current state-of-the-art computational  methods~\cite{zhang2016single,ma2019bayesian,m_Wang-etal-NIPS20} for counting can only handle a limited number of visual categories. In fact, most of the counting neural networks~\cite{arteta2016counting,m_Wang-etal-NIPS20} can handle a single category at a time, such as people, cars, and cells. 

There are two major challenges preventing the Computer Vision community from designing systems capable of counting a large number of visual categories. First, most of the contemporary counting approaches~\cite{m_Wang-etal-NIPS20,zhang2016single,arteta2016counting} treat counting as a supervised regression task, requiring thousands of labeled images to learn a fully convolutional regressor that maps an input image to its corresponding density map, from which the estimated count is obtained by summing all the density values. These networks require dot annotations for millions of objects on several thousands of training images, and obtaining this type of annotation is a costly and laborious process. As a result, it is difficult to scale these contemporary counting approaches to handle a large number of visual categories. Second, there are not any large enough unconstrained counting datasets with many visual categories for the development of a general counting method. Most of the popular counting datasets~\cite{zhang2016single,Idrees_2013_CVPR,idrees2018composition,wang2020nwpu,sindagi2020jhu,hsieh2017drone} consist of a single object category.

%. As shown in \Tref{tab:DatasetComparison}, most of the popular counting datasets consist of a single visual category.
\begin{figure}[t]
     \includegraphics[width=\linewidth]{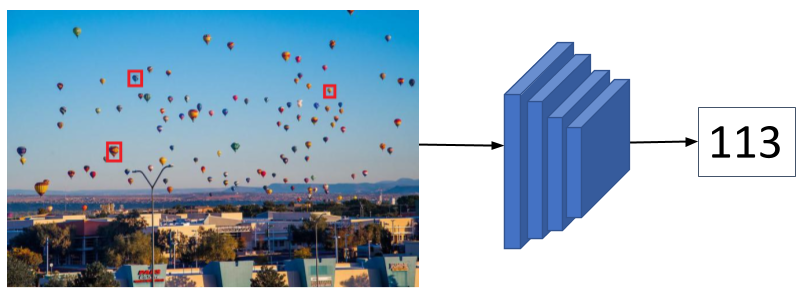}
     \vskip -0.1in
    \caption{\textbf{Few-shot counting---the objective of our work.} Given an image from a novel class and a few exemplar objects from the same image delineated by bounding boxes, the objective is to count the total number of objects of the novel class in the image.
  \label{fig:motivation}}
 \end{figure} 

\iffalse
\setlength{\tabcolsep}{3pt}
\begin{table}[!tb]
\centering
\begin{tabular}{lcccc}
\toprule
& & & \multicolumn{2}{c}{Annotation type} \\
\cmidrule(lr){4-5}
Dataset & Images & Categories & Dot & Bounding Box  \\
\midrule
UCF CC 50~\cite{Idrees_2013_CVPR} & 50  & 1 & \cmark & \xmark  \\
Shanghaitech~\cite{zhang2016single}  & 1198 & 1  & \cmark & \xmark  \\
UCF QNRF~\cite{idrees2018composition} & 1535 & 1   & \cmark & \xmark    \\
NWPU~\cite{wang2020nwpu} & 5109 & 1& \cmark & \xmark \\
JHU Crowd~\cite{sindagi2020jhu} & 4372 & 1  & \cmark & \cmark \\
CARPK~\cite{hsieh2017drone} & 1448 & 1  & \cmark & \cmark   \\
\textbf{Proposed} & 6135 & 147 & \cmark & \cmark   \\
\bottomrule
\end{tabular}
%\vskip 0.15in
\caption{{\bf Comparison with popular counting datasets}. Existing  datasets comprise of a single object category, while our dataset contain multiple categories. Our dataset is of similar size as the existing datasets, and it comes with dot, bounding box, and polygon annotations. \mhoai{This table is not needed, at least not in the introduction section}
\label{tab:DatasetComparison}}
\end{table}
\fi

In this work, we address both of the above challenges. To handle the first challenge, we take a detour from the existing counting approaches which treat counting as a typical fully supervised regression task, and pose counting as a few shot regression task, as shown in \Fref{fig:motivation}. In this few-shot setting, the inputs for the counting task are an image and few examples from the same image for the object of interest, and the output is the count of object instances. The examples are provided in the form of bounding boxes around the objects of interest. In other words, our few shot counting task deals with counting instances within an image which are similar to the exemplars from the same image. Following the convention from the few-shot classification task~\cite{Lake-et-al-Science15,vinyals2016matching,finn2017model}, the classes at test time are completely different from the ones seen during training. This makes few-shot counting very different from the typical counting task, where the training and test classes are the same. Unlike the typical counting task, where hundreds~\cite{zhang2016single} or thousands~\cite{idrees2018composition} of labeled examples are available for training, a few-shot counting method needs to generalize to completely novel classes using only the input image and a few exemplars. 

% \vrDone{M2N might not be the best name. Based on the name the emphasize is non Multi-scale. What you want to emphasize is the two parts: few-shot matching + adaptation. So you need to create a name from Exemplar or Few Shot, with Matching and Adaptation. ExeMplar Matching and Adaptation Network (FamNet)? }

We propose a novel architecture called \underline{F}ew Shot \underline{A}daptation and \underline{M}atching Network~(FamNet) for tackling the few-shot counting task. FamNet has two key components: 1) a feature extraction module, and 2) a density prediction module. The feature extraction module consists of a general feature extractor capable of handling a large number of visual categories. The density prediction module is designed to be agnostic to the visual category. As will be seen in our experiments, both the feature extractor and density prediction modules can already generalize to the novel categories at test time. We further improve the performance of FamNet by developing a novel few-shot adaptation scheme at test time. This adaptation scheme uses the provided exemplars themselves and adapts the counting network to them with a few gradient descent updates, where the gradients are computed based on two loss functions which are designed to utilize the locations of the exemplars to the fullest extent. Empirically, this adaptation scheme improves the performance of FamNet.

%utilize their locations to the fullest extent. 

%\vrDone{Viresh, move the part on few-shot classification to related work}
%Many of the few-shot classification approaches~\cite{finn2017model,ravi2016optimization} adapt the network to the novel categories at test time, resulting in improved performance. Drawing inspiration from these works, we propose a novel adaptation scheme which utilizes the exemplars available at test time and performs a few steps of gradient descent in order to adapt FamNet to any novel category. We propose two loss functions which can utilize the exemplar information for the adaptation. We show that the adaptation scheme helps us in improving the performance of FamNet.

Finally, to address the lack of a dataset for developing and evaluating the performance of few-shot counting methods, we introduce a medium-scale dataset consisting of more than 6000 images from 147 visual categories. The dataset comes with dot and bounding box annotations, and is suitable for the few-shot counting task. We name this dataset Few-Shot Counting-147 (FSC-147).

%There does not exist any unconstrained counting dataset comprising of large number of visual categories suitable for the few-shot counting task. Hence, we collect a new dataset called few-shot Counting-147(FSC-147), comprising of 147 visual categories and over 6k images\footnote{the dataset and code will be made public post acceptance.}. The dataset comes with dot annotations and bounding box annotations, and is suitable for the few-shot counting task outlined above.

In short, the main contributions of our work are as follows. First, we pose counting as a few-shot regression task. Second, we propose a novel architecture called FamNet for handling the few-shot counting task, with a novel few-shot adaptation scheme at test time. Third, we present a novel few-shot counting dataset called FSC-147, comprising of over 6000 images with 147 visual categories.

% We also collect a visual counting dataset consisting of significantly more object classes than the existing datasets. We contrast our dataset with the existing ones in \Tref{tab:DatasetComparison}.

\section{Related Works}
In this work, we are interested in counting objects of interest in a given image with a few labeled examples from the same image. Most of the previous counting methods are for specific types of objects such as people~\cite{ma2019bayesian,zhang2016single,ranjan2018iterative,babu2018divide,sam2017switching,li2018csrnet,liu2018leveraging,cao2018scale,ranjan2019crowd,shi2019revisiting,liu2019context,wang2019learning,zhang2019attentional,wan2019adaptive,m_Ranjan-etal-ACCV20,m_Abousamra-etal-AAAI21}, cars~\cite{mundhenk2016large}, animals~\cite{arteta2016counting}, cells~\cite{arteta2016detecting,xie2018microscopy,khan2016deep}, and fruits~\cite{rahnemoonfar2017deep}. These methods often require training images with tens of thousands or even millions of annotated object instances. Some of these works~\cite{m_Ranjan-etal-ACCV20} tackle the issue of costly annotation cost to some extent by adapting a counting network trained on a source domain to any target domain using labels for only few informative samples from the target domain. However, even these approaches require a large amount of labeled data in the source domain.  

The proposed FamNet works by exploiting the strong similarity between a query image and the provided exemplar objects in the image. To some extent, it is similar the decade-old self-similarity work of Shechtman and Irani~\cite{Shechtman-Irani-CVPR07}. Also related to this idea is the recent work of Lu and Zisserman\cite{lu2018class}, who proposed a Generic Matching Network (GMN) for class-agnostic counting. GMN was pre-trained with tracking video data, and it had an explicit adaptation module to adapt the network to an image domain of interest. GMN has been shown to work well if several dozens to hundreds of examples are available for adaptation. Without adaptation, GMN does not perform very well on novel classes, as will be seen in our experiments. 

%
%Some of the recent works~\cite{lu2018class} have focussed on class agnostic counting.
%Lu  ~\cite{lu2018class} present a two stage class agnostic counting approach, where the first stage consists of training a generic matching network on video frames. The second stage consists of adapting the matching network on any counting dataset. unlike ~\cite{lu2018class}, our proposed approach can be directly trained on any novel class using a few labeled examples, doing away with the need for any expensive matching network training.    

Related to few-shot counting is the few-shot detection task (e.g.,~\cite{kang2019few,fan2020few}), where the objective is to learn a detector for a novel category using a few labeled examples. Few-shot counting differs from few-shot detection in two primary aspects. First, few-shot counting requires dot annotations while detection requires bounding box annotations. Second, few-shot detection methods can be affected by severe occlusion whereas few-shot counting is tackled with a density estimation approach~\cite{Lempitsky-Zisserman-NIPS10,zhang2016single}, which is more robust towards occlusion than the detection-then-counting approach because the density estimation methods do not have to commit to binarized decisions at an early stage. The benefits of the density estimation approach has been empirically demonstrated in several domains, especially for crowd and cell counting.

Also related to our work is the task of few-shot image classification~\cite{lake2015human,koch2015siamese,vinyals2016matching,santoro2016one,finn2017model,ravi2016optimization}. The few-shot classification task deals with classifying images from novel categories at test time, given a few training examples from these novel test categories.
The Model Agnostic Meta Learning (MAML)~\cite{finn2017model} based few-shot approach is relevant for our few-shot counting task, and it focuses on learning parameters which can adapt to novel classes at test time by means of few gradient descent steps. However, MAML involves computing second order derivatives during training which makes it expensive, even more so for the pixel level prediction task of density map prediction being considered in our paper. Drawing inspiration from these works, we propose a novel adaptation scheme which utilizes the exemplars available at test time and performs a few steps of gradient descent in order to adapt FamNet to any novel category. Unlike MAML, our training scheme does not require higher order gradients at training time. We compare our approach with MAML, and empirically show that it leads to better performance and is also much faster to train.

\begin{figure*}[t]
    \includegraphics[width=\textwidth]{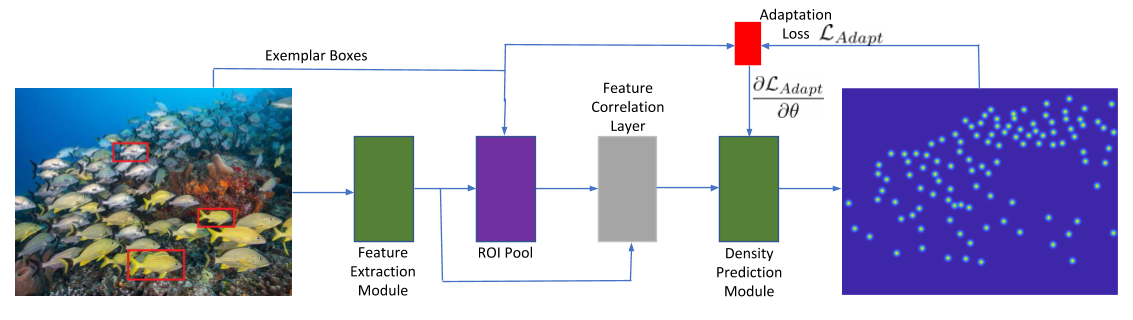}
    \caption{{\bf Few-shot adaptation \& matching Network} takes as input the query image along with few bounding boxes depicting the object of interest, and predicts the density map. The count is obtained by summing all the pixel values in the density map. The adaptation loss is computed based on the bounding box information, and the gradients from this loss are used to update the parameters of the density prediction module. The adaptation loss is only used during test time. 
  \label{fig:EmNet}}
 \end{figure*}  
\section{Few-Shot Adaptation \& Matching Network}

In this section, we describe the proposed FamNet for tackling the few-shot counting task. 

\subsection{Network architecture}

\Fref{fig:EmNet} depicts the pipeline of FamNet. The input to the network is an image $X \in \Re^{H \times W \times 3}$ and a few exemplar bounding boxes depicting the object to be counted from the same image. The output of the network is the predicted density map $Z \in \Re^{H \times W}$, and the count for the object of interest is obtained by summing over all density values. 
 
FamNet consists of two key modules: 1) a multi-scale feature extraction module, and 2) a density prediction module. We design both of these modules so that they can handle novel categories at test time. We use an ImageNet-pretrained network~\cite{He-et-al-CVPR16} for the feature extraction, since such networks can handle a broad range of visual categories. The density prediction module is designed to be agnostic to the visual categories.
The multi-scale feature extraction module consists of the first four blocks from a pre-trained ResNet-50 backbone~\cite{He-et-al-CVPR16} (the parameters of these blocks are frozen during training). We represent an image by the convolutional feature maps at the third and fourth blocks. We also obtain the multi-scale features for an exemplar by performing ROI pooling on the convolutional feature maps from the third and fourth Resnet-50 blocks.

To make the density prediction module agnostic to the visual categories, we do not use the features obtained from the feature extraction module directly for  density prediction. Instead, we only use the correlation map between the exemplar features and image features as the input to the density prediction module. To account for the objects of interest at different scales, we scale the exemplar features to different scales, and correlate the scaled exemplar features with the image features to obtain multiple correlation maps, one for each scale. For all of our experiments, we use the scales of 0.9 and 1.1, along with the original scale. The correlation maps are concatenated and fed into the density prediction module. The density prediction module consists of five convolution blocks and three upsampling layers placed after the first, second, and third convolution layers. The last layer is a $1{\times}1$ convolution layer, which predicts the 2D density map. The size of the predicted density map is the same as the size of the input image. 

\subsection{Training}

We train the FamNet using the training images of our dataset. Each training image contains multiple objects of interest, but only the exemplar objects are annotated with bounding boxes and the majority of the objects only have dot annotations. It is, however, difficult to train a density estimation network with the training loss that is defined based on the dot annotations directly. Most existing works for visual counting, especially for crowd counting~\cite{zhang2016single}, convolve the dot annotation map with a Gaussian window of a fixed size, typically $15{\times}15$, to generate a smoothed target density map for training the density estimation network. 

Our dataset consists of $147$ different categories, where there is huge variation in the sizes of the objects. Therefore, to generate the target density map, we use Gaussian smoothing with adaptive window size. First, we use dot annotations to estimate the size of the objects. Given the dot annotation map, where each dot is at an approximate center of an object, we compute the distance between each dot and its nearest neighbor, and average these distances for all the dots in the image. This average distance is used as the size of the Gaussian window to generate the target density map. The standard deviation of the Gaussian is set to be a quarter of the window size. 

To train FamNet, we minimize the mean squared error between the predicted density map and the ground truth density map. We use Adam optimizer with a learning rate of $10^{-5}$, and batch size of 1. We resize each image to a fixed height of 384, and the width is adjusted accordingly to preserve the aspect ratio of the original image. 
% \begin{align}
%     \mL_{Train} = \frac{1}{K}\sum_{k=1}^{K} || Z^k - G^k||_2^2.
% \end{align}
% \mhoai{Do you minimize the mean squared error or sum of squared error? Does the above equation must be normalized by the size of the image?}
%There are 697 training images whose resized width is less than 384. Such images are resized to $384{\times}408$ before training.  
%Let's not shoot ourselves in the foot here.

\subsection{Test-time adaptation}

Since the two modules of the FamNet are not dependent on any object categories, the trained FamNet can already be used for counting objects from novel categories given a few exemplars. In this section, we describe a novel approach to adapt this network to the exemplars, further improving the accuracy of the estimated count. The key idea is to harness the information provided by the locations of the exemplar bounding boxes. So far, we have only used the bounding boxes of the exemplars to extract appearance features of the exemplars, and we have not utilized their locations to the full extent. 

%That being said, the exemplar bounding box locations available at the test time can be used to adapt the M2N to the test category corresponding to the object of interest.

Let $B$ denote the set of provided exemplar bounding boxes. For a bounding box $b \in B$, let $Z_b$ be the crop from the density map $Z$ at location $b$. To harness the extra information provided by the locations of the bounding boxes $B$, we propose to consider the following two losses.

\myheading{Min-Count Loss.} For each exemplar bounding box~$b$, the sum of the density values within $Z_b$ should be at least one. This is because the predicted count is taken as the sum of predicted density values, and there is at least one object at the location  specified by the bounding box $b$. However, we cannot assert that the sum of the density values within $Z_b$ to be exactly one, due to possible overlapping between $b$ and other nearby objects of interest. This observation leads to an inequality constraint: $||Z_b||_1 \geq 1$, where $||Z_b||_1$ denotes the sum of all the values in $Z_b$. Given the predicted density map and the set of provided bounding boxes for the exemplars, we define the following Min-Count loss to quantify the amount of constraint violation: 

\begin{equation}\label{adaptation1}
\mL_{MinCount} = \sum_{b \in B} \max(0,1- ||Z_b||_1).    
\end{equation}

% The scalar value at any location in the ground truth map might be resulting from summing values from different Gaussian kernels corresponding to nearby objects. At test time, only a single/few bounding boxes are known, and the possible nearby bounding boxes are unknown. Hence, it is not possible to obtain the ground truth density values within the bounding box based on the location of the bounding box. However, we do know that the ground truth value within bounding box should be at least 1. We use this knowledge and define a margin loss as:
% \begin{equation}\label{adaptation1}
% \mL_{margin} = \sum_{b \in B} \max(0,1- ||Z_b||_1).    
% \end{equation}
% where $||Z_b||_1$ refers to the sum of all the pixels from the predicted density map lying within the bounding box $b$. This loss results in a non-zero value if the sum of all the pixels within the bounding box is less than 1.

\myheading{Perturbation Loss.} Our second loss to harness the positional information provided by the exemplar bounding boxes is inspired by the success of tracking algorithms based on correlation filter \cite{Henriques-etal-PAMI15,valmadre2017end,wang2019fast}. Given the bounding box of an object to track, these algorithms learn a filter that has highest response at the exact location of the bounding box and lower responses  at perturbed locations. The correlation filter can be learned by  optimizing a regression function to map from a perturbed location to a target response value, where the target response value decreases exponentially as the perturbation distance increases, usually specified by a Gaussian distribution. 

In our case, the predicted density map $Z$ is essentially the correlation response map between the exemplars and the image. To this end, the density values around the location of an exemplar should ideally look like a Gaussian. Let $G_{h \times w}$ be the 2D Gaussian window of size $h{\times}w$. We define the perturbation loss as follows: 
\begin{equation}\label{adaptation2}
\mL_{Per} = \sum_{b \in B} ||Z_b - G_{h \times w}||_2^2.    
\end{equation}

% The margin loss imposes constraints on the sum of all the pixels within the predicted density map $Z_i$ which lie within the bounding box $B_i$. However, the margin loss doesn't impose any strict pixel level constraints, which may end up making the prediction less sharp. Furthermore, the margin loss won't impose any constraints if the sum of pixels within the bounding box is more than 1. Hence, we use a 2D Gaussian kernel with size equal to the size of the bounding box as an approximation to the groundtruth within the bounding box to impose pixel level constraints. Let the Gaussian be represented by $G_{h \times w} \in \Re^{h \times w}$ where $h$ and $w$ are the height and width of the bounding box. We define the pixel wise Gaussian loss as follows
% \begin{equation}\label{adaptation2}
% Gaussian \hspace{0.2cm} loss = MSE(Z^{B}, \hspace{0.1cm} G_{h \times w})    
% \end{equation}
% where MSE refers to the mean squared loss function.

\myheading{The combined adaptation Loss.} The loss used for test-time adaptation is the weighted combination of the Min-Count loss and the Perturbation loss. The final test time adaptation loss is given as 
\begin{equation}\label{adaptation3}
    \mL_{Adapt} = \lambda_1 \mL_{MinCount} + \lambda_2 \mL_{Per},
\end{equation}
where $\lambda_1$ and $\lambda_2$ are scalar hyper parameters. At test time, we perform $100$ gradient descent steps for each test image, and optimize the joint loss presented in  \Eref{adaptation3}. We use the learning rate $10^{-7}$. The values for $\lambda_1$ and $\lambda_2$ are $10^{-9}$ and $10^{-4}$ respectively. 
The learning rate, the number of gradient steps, $\lambda_1$, and $\lambda_2$, are tuned based on the performance on the validation set. 
The values of $\lambda_1$, and $\lambda_2$ seem small, but this is necessary to make the adaptation loss to have similar magnitude to the training loss. Even though the training loss is not used for test time adaptation, it is important for the losses and their gradients to have similar magnitudes. Otherwise, the gradient update steps of the adaptation process will either do nothing or move away far from the parameters learned during training.

% \mhoai{Why are the values o $\lambda_1$ and $\lambda_2$ so small? What is the learning rate for gradient descent? One can achieve equivalent formulation by scaling up $\lambda_1$ and $\lambda_2$ while scaling down the learning rate. }
% \vr{Original training loss are divided by 384*384 (mse loss divided by output size). Perturbation loss is computed over roughly 20*20 windows, hence divided by 20*20. They have larger magnitude than training loss. Mincount loss also has large magnitude than training losses.That's why small lambda values}

% \mhoai{The two losses are not quite compatible in terms of types. $\mL_{MinCount}$ is based on the margin loss of $L_1$ norm, while $\mL_{Per}$ is based on the Squared $L_2$ norm. This would make it difficult to combine and tune them. This seems to be a sub-optimal design. } 
% \vr{$\lambda_1$ and $\lambda_2$ are chosen so that the order of magnitude of both losses becomes same/similar} 

Note that the adaptation loss is only used at test time. During training of FamNet, this loss is redundant because the proposed training loss, based on mean squared errors computed over all pixel locations, already provides stronger supervision signal than the adaptation loss.

\section{The FSC-147 Dataset}
\begin{figure*}[!t]
    \begin{subfigure}[!category]{0.3\textwidth}
    \includegraphics[height=1\textwidth, width=\textwidth]{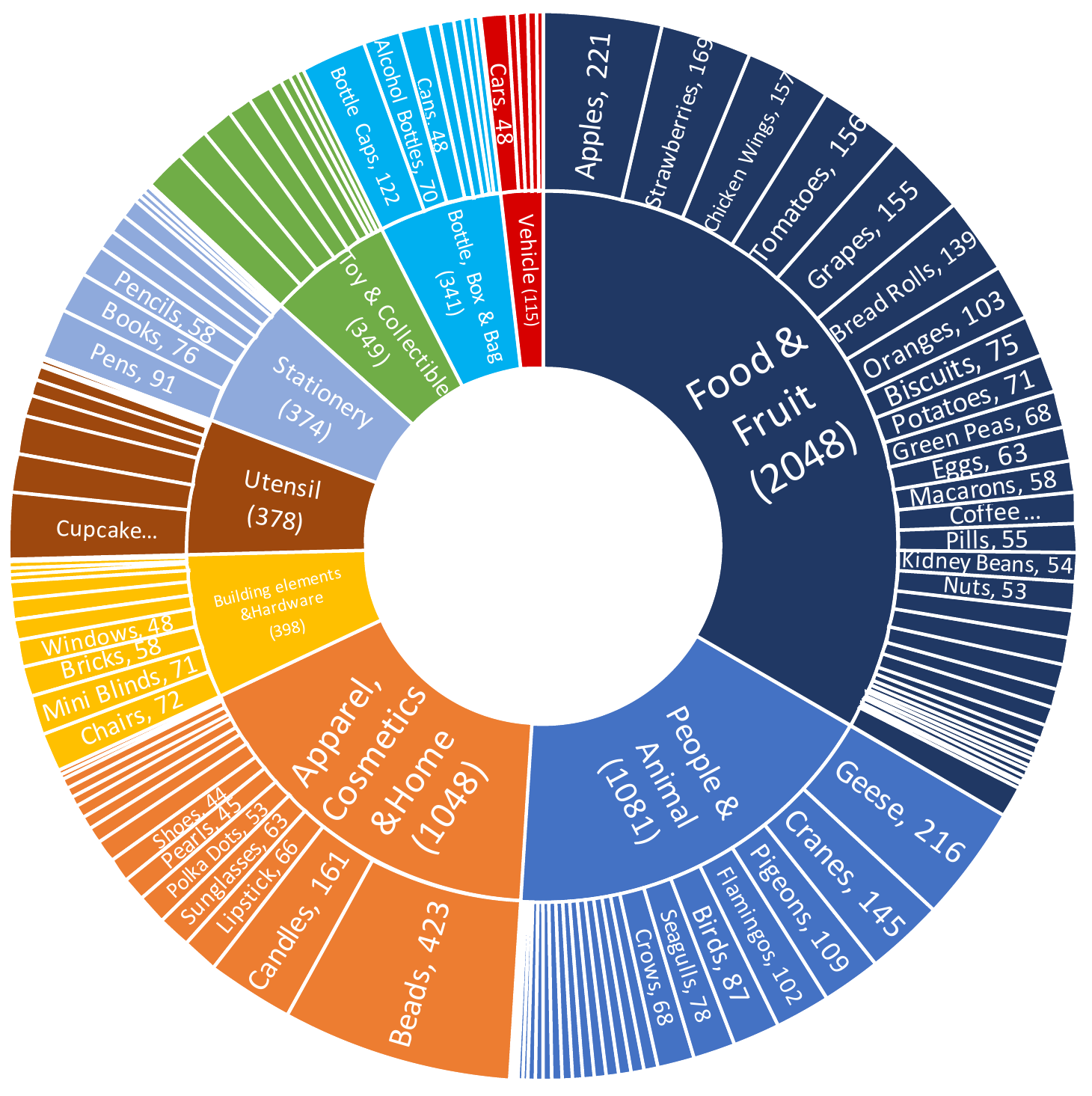}
    \label{fig:cate}
    \vskip -0.1in
    \caption[width=0.8\textwidth]{Image categories and number of images for each category in our dataset. \label{fig:categoryPlot}}
\end{subfigure}
    \hfill
    \begin{subfigure}[!hist]{0.3\textwidth}
        \includegraphics[height=\textwidth, width=\textwidth]{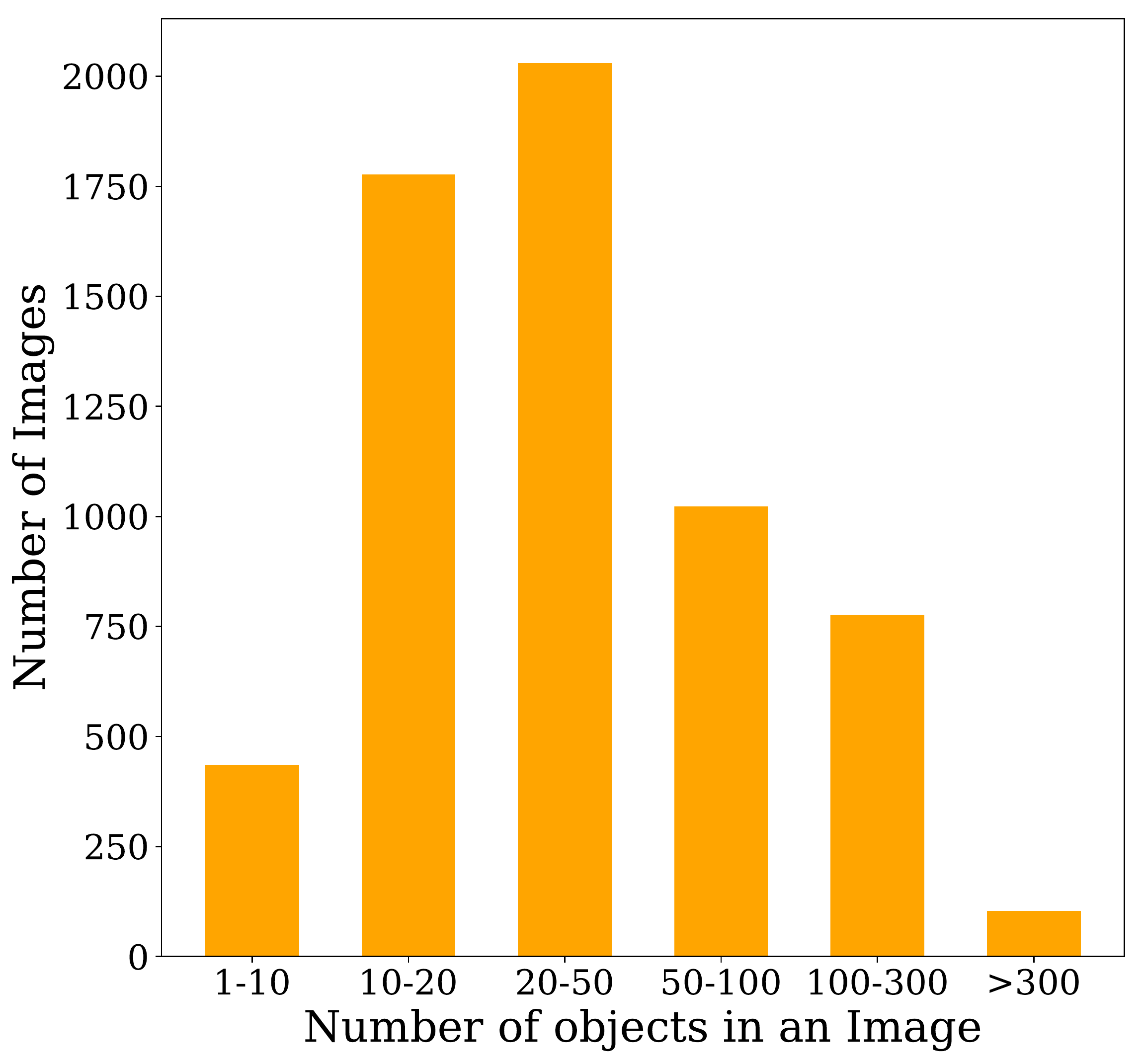}
        \label{fig:hist}
    \vskip -0.1in        
        \caption{Number of images in several ranges of object count.\label{fig:histogram}}
    \end{subfigure}
     \hfill
\begin{subfigure}[!hist]{0.37\textwidth}
\setlength{\tabcolsep}{2pt}
\begin{adjustbox}{width=\columnwidth}
\begin{tabular}{lcccc}
\toprule
& & & \multicolumn{2}{c}{Annotation type} \\
\cmidrule(lr){4-5}
Dataset & Images & Categories & Dot & Bounding Box  \\
\midrule
UCF CC 50~\cite{Idrees_2013_CVPR} & 50  & 1 & \cmark & \xmark  \\
Shanghaitech~\cite{zhang2016single}  & 1198 & 1  & \cmark & \xmark  \\
UCF QNRF~\cite{idrees2018composition} & 1535 & 1   & \cmark & \xmark    \\
NWPU~\cite{wang2020nwpu} & 5109 & 1& \cmark & \xmark \\
JHU Crowd~\cite{sindagi2020jhu} & 4372 & 1  & \cmark & \cmark \\
CARPK~\cite{hsieh2017drone} & 1448 & 1  & \cmark & \cmark   \\
\textbf{Proposed} & 6135 & 147 & \cmark & \cmark   \\
\bottomrule
\end{tabular}
\end{adjustbox}
\vskip .1in
\caption{Comparison with popular counting datasets. 
\label{tab:DatasetComparison}}
    \end{subfigure}
 \label{fig:plot}
  \caption{{\bf Categories \& no. of images per category, object counts, and comparison with other counting datasets}}
 \end{figure*}  

To train the FamNet, we need a dataset suitable for the few-shot counting task, consisting of many visual categories. Unfortunately, existing counting datasets are mostly dedicated for specific object categories such as people, cars, and cells. Meanwhile, existing multi-class datasets do not contain many images that are suitable for visual counting. For example, although some images from the COCO dataset~\cite{Lin-etal-ECCV14} contains multiple instances from the same object category, most of the images do not satisfy the conditions of our intended applications due to the small number of object instances or the huge variation in pose and appearance of the object instances in each image. 

Since there was no dataset that was large and diverse enough for our purpose, we collected and annotated images ourselves. Our dataset consists of 6135 images across a diverse set of 147 object categories, from kitchen utensils and office stationery to vehicles and animals. The object count in our dataset varies widely, from 7 to 3731 objects, with an average count of 56 objects per image. In each image, each object instance is annotated with a dot at its approximate center. In addition, three object instances are selected randomly as exemplar instances; these exemplars are also annotated with axis-aligned bounding boxes. In the following subsections, we will describe how the data was collected and annotated. We will also report the detailed statistics and how the data was split into disjoint training, validation, and testing sets.

%

%We propose a large scale class agnostic counting dataset, Visual Counting 64 (VC64) which has in total 1005 images across 64 categories. These categories span a diverse set ranging from Animals, Cars, Cutlery to Clothes, Office Supplies etc touching every aspect of our day to day life. Table \ref{tab:dataset_categories} contains few categories of VC64 and their images count.

\subsection{Image Collection}

To obtain the set of 6135 images for our dataset, we started with a set of candidate images obtained by keyword searches. Subsequently, we performed manual inspection to filter out images that do not satisfy our predefined conditions as described below.

\myheading{Image retrieval}. We started with a list of object categories, and collected 300--3000 candidate images for each category by scraping the web. We used Flickr, Google, and Bing search engines with the open source image scrappers~\cite{Flickr-Scrapper,Google-Scrapper}. We added adjectives such as \textit{many, multiple, lots of}, and \textit{stack of} in front of the category names to create the search query keywords.

\myheading{Manual verification and filtering}. 
We manually inspected the candidate images and only kept the suitable ones satisfying the following criteria: 
\begin{enumerate} \denselist
    \item \textit{High image quality}: The resolution should be high enough to easily differentiate between objects. 
    \item \textit{Large enough object count}: The number of objects of interest should be at least 7. We are more interested in counting a large number of objects, since humans do not need help counting a small number of objects. 
    \item \textit{Appearance similarity}: we selected images where object instances have somewhat similar poses, texture, and appearance.  
    \item \textit{No severe occlusion}: in most cases, we removed candidate images where severe occlusion prevents humans from accurately counting the objects. 
\end{enumerate}

\subsection{Image Annotation}
\begin{figure*}[!th]

  \begin{subfigure}[b]{0.23\textwidth}    \includegraphics[height=0.75\textwidth,width=\textwidth]{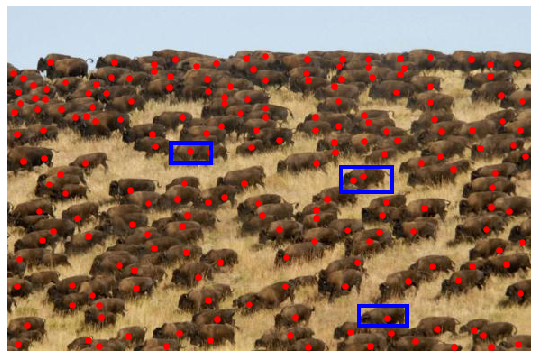}
    % \caption{Dot Annotation.}
    % \label{fig:chairs_dot}
  \end{subfigure}
  \hfill
  \begin{subfigure}[b]{0.23\textwidth}    \includegraphics[height=0.75\textwidth,width=\textwidth]{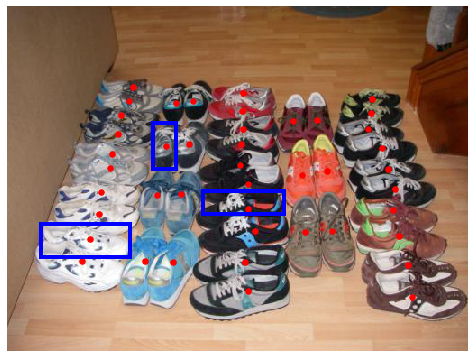}
    % \caption{Dot Annotation.}
    % \label{fig:chairs_dot}
  \end{subfigure}
  \hfill
    \begin{subfigure}[b]{0.23\textwidth}    \includegraphics[height=0.75\textwidth,width=\textwidth]{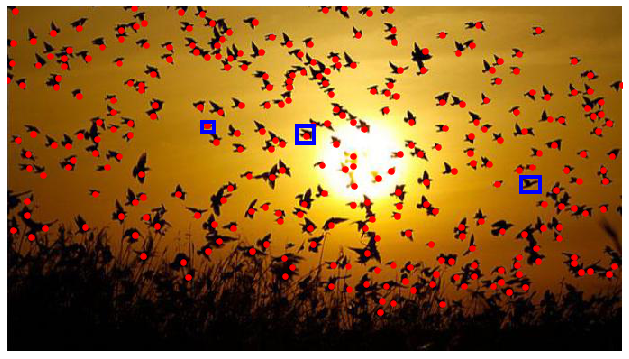}%1866 {./Images/examples/4077.png}
    % \caption{Dot Annotation.}
    % \label{fig:chairs_dot}
  \end{subfigure}
  \hfill
      \begin{subfigure}[b]{0.23\textwidth}    \includegraphics[height=0.75\textwidth,width=\textwidth]{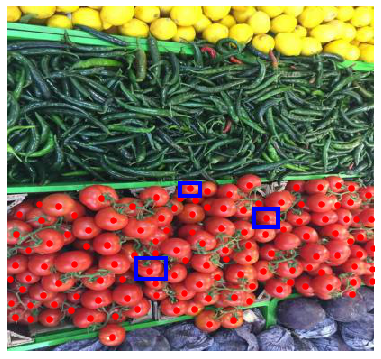}
    % \caption{Dot Annotation.}
    % \label{fig:chairs_dot}
  \end{subfigure}
  \vspace{3pt}

  \begin{subfigure}[b]{0.23\textwidth}    \includegraphics[height=0.75\textwidth,width=\textwidth]{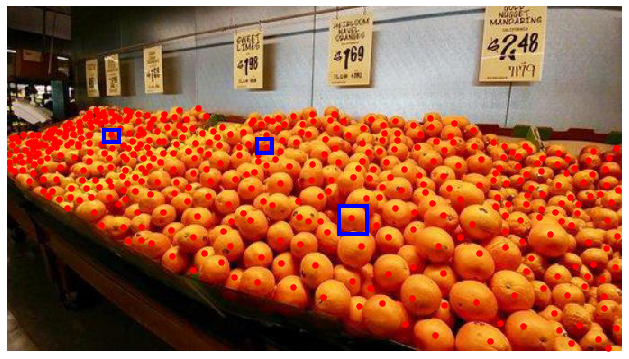}
    % \caption{Dot Annotation.}
    % \label{fig:chairs_dot}
  \end{subfigure}
  \hfill
  \begin{subfigure}[b]{0.23\textwidth}    \includegraphics[height=0.75\textwidth,width=\textwidth]{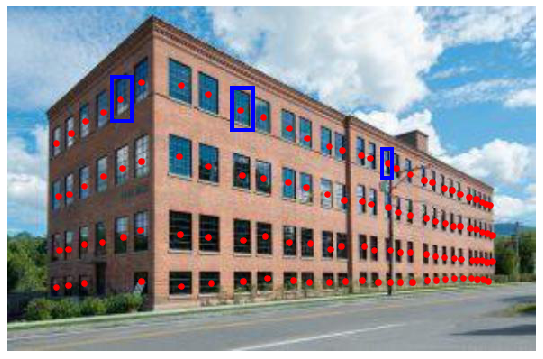}
    % \caption{Dot Annotation.}
    % \label{fig:chairs_dot}
  \end{subfigure}
  \hfill
    \begin{subfigure}[b]{0.23\textwidth}    \includegraphics[height=0.75\textwidth,width=\textwidth]{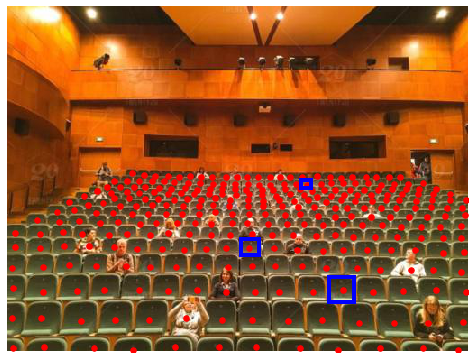}
    % \caption{Dot Annotation.}
    % \label{fig:chairs_dot}
  \end{subfigure}
  \hfill
      \begin{subfigure}[b]{0.23\textwidth}    \includegraphics[height=0.75\textwidth,width=\textwidth]{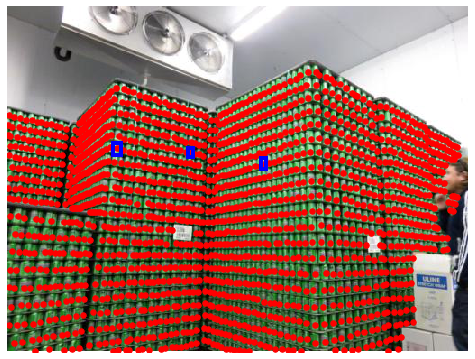}%2686.png}
    % \caption{Dot Annotation.}
    % \label{fig:chairs_dot}
  \end{subfigure}
%   \vspace{3pt}
\vskip -0.05in
  \caption{{\bf Few annotated images from the dataset}. Dot and box annotations are shown in red and blue respectively. The number of objects in each image varies widely, some images contain a dozen of objects while some contains thousands.  \label{fig:annotation_examples} }

\end{figure*}

Images in the dataset were annotated by a group of annotators using the OpenCV Image and Video Annotation Tool~\cite{CVAT}. Two types of annotation were collected for each image, dots and bounding boxes, as illustrated in  \Fref{fig:annotation_examples}. For images containing multiple categories, we picked only one of the categories. Each object instance in an image was marked with a dot at its approximate center. In case of occlusion, the occluded instance was only counted and annotated if the amount of occlusion was less than 90\%. For each image, we arbitrarily chose three objects as exemplar instances and we drew axis-aligned bounding boxes for those instances. 

\subsection{Dataset split}
We divided the dataset into train, validation, and test sets such that they do not share any object category. We randomly selected 89 object categories for the train set, and 29 categories each for the validation and test sets. The train, validation, and test sets consist of 3659, 1286 and 1190 images respectively.
 
\subsection{Data Statistics}
\iffalse
\vr{The statistics need to be changed in this section, text will remain unchanged.}
The dataset contains images from 148 categories. These categories can be roughly grouped to form super-categories: (i) People \& Animal (156 images); (ii) Apparel \& Cosmetic (133 images); (iii) Bottle, Box, \& Bag (129 images); (iv) Furniture and building elements (84 images); (v) Food \& Fruit (130 images); (vi) Toy \& Collectible (84 images); (vii) Office stationery (164 images); and (viii) Vehicle (49 images). \Fref{fig:categoryPlot} shows the categories and the super-categories together with the number of images for each category. 
\fi

The dataset contains a total of 6135 images. The average height and width of the images are 774 and 938 pixels, respectively. The average number of objects per image is 56, and the total number of objects is \mbox{343,818}. The minimum and maximum number of objects for one image are 7 and 3701, respectively. The three categories with the highest number of objects per image are: Lego (303 objects/image), Brick (271), and Marker (247). The three categories with lowest number of objects per image are: Supermarket shelf (8 objects/image), Meat Skewer (8), and Oyster (11). \Fref{fig:histogram} is a histogram plot for the number of images in several ranges of object count.

\section{Experiments}

\subsection{Performance Evaluation Metrics} 

We use Mean Absolute Error (MAE) and Root Mean Squared Error (RMSE) to measure the accuracy of a counting method. MAE and RMSE are commonly used metrics for counting task~\cite{zhang2016single,ma2019bayesian,ranjan2018iterative}, and they are defined as follows.
$MAE = \frac{1}{n}\sum_{i=1}^{n} \lvert c_i - \hat{c}_i \rvert; 
RMSE = \sqrt[]{\frac{1}{n}\sum_{i=1}^{n} (c_i - \hat{c}_i)^2}, 
$
where $n$ is the number of test images, and $c_i$ and $\hat{c}_i$ are the ground truth and predicted counts.

\subsection{Comparison with Few-Shot Approaches}
\setlength{\tabcolsep}{2pt}
\begin{table}[!tb]
\vskip 0.1in
\centering
\begin{tabular}{lcccc}
\toprule
          &  \multicolumn{2}{c}{ Val Set} &  \multicolumn{2}{c}{ Test Set} \\
         \cmidrule(lr){2-3} \cmidrule(lr){4-5} 
    Method &  MAE          & RMSE    & MAE          & RMSE         \\

\midrule 
Mean & 53.38 & 124.53 & 47.55 & 147.67 \\
Median & 48.68 & 129.70 & 47.73 & 152.46 \\
%{\small pre-trained few-shot detector}\cite{kang2019few} & 50.76 & 126.18 & 44.16 & 146.76 \\
FR few-shot detector  \cite{kang2019few} & 45.45  & 112.53 & 41.64 & 141.04 \\
FSOD few-shot detector \cite{fan2020few} & 36.36 & 115.00 & 32.53  & 140.65 \\
Pre-trained GMN \cite{lu2018class} & 60.56 &  137.78 & 62.69 & 159.67 \\
GMN \cite{lu2018class} & 29.66 & 89.81 & 26.52 & 124.57  \\
MAML \cite{finn2017model} & 25.54 & 79.44 & 24.90  & 112.68 \\
%\color{blue}{
%MAML} \cite{finn2017model} & 29.55 & 87.59 & 26.84  & 110.98 \\
%FamNet (Proposed)   & \textbf{26.80} & \textbf{73.83} & \textbf{24.63} & \textbf{102.28}  \\
%FamNet (Proposed)}   & \textbf{24.97} & \textbf{79.34} & \textbf{21.87} & \textbf{100.22}  \\
FamNet (Proposed)   & \textbf{23.75} & \textbf{69.07} & \textbf{22.08} & \textbf{99.54}  \\
%FamNet (Proposed)   & \textbf{22.76} & \textbf{75.81} & \textbf{21.77} & \textbf{108.90}  \\
\bottomrule 
\end{tabular}
\vskip -0.1in
\caption{Comparing FamNet to two simple baselines (Mean, Median) and four stronger baseline (Feature Reweighting (FR) few-shot detector, FSOD few-shot detector, GMN and MAML), these are few-shot methods that have been adapted and trained for counting. FamNet has the lowest MAE and RMSE on both val and test sets.
\label{tab:baseline}}
\end{table}

We compare the performance of FamNet with two trivial baselines and four competing few-shot methods. The two trivial baseline methods are: (1) always output the average object count for training images; (2) always output the median count for the training images. We also implement stronger methods for comparison, by adapting several few-shot methods for the counting task and training them on our training data. Specifically, we adapt the following approaches for counting: the state-of-the-art few-shot detectors~\cite{kang2019few,fan2020few}, the Generic Matching Network (GMN)~\cite{lu2018class}, and Model Agnostic Meta Learning (MAML)~\cite{finn2017model}. 
We implement MAML using the higher library~\cite{grefenstette2019generalized}, which is a meta learning library supporting higher order optimization. The training procedure of MAML involves an \textit{inner optimization loop}, which adapts the network to the specific test classes, and an \textit{outer optimization loop} which learns meta parameters that facilitate faster generalization to novel tasks. At test time, only the inner optimization is performed. We use the $\mL_{Adapt}$ loss defined in \Eref{adaptation3} for the inner optimization loop, and the MSE loss over the entire dot annotation map for the outer optimization loop. 
%Note that GMN uses significantly more training data than all the other approaches given its pre-training stage which uses ILSVRC video dataset. None of the other approaches use the ILSVRC video data. 

As can be seen in \Tref{tab:baseline}, FamNet outperforms all the other methods. Surprisingly, the pre-trained GMN does not work very well, even though it is a class agnostic counting method. The GMN model trained on our training data performs better than its pre-trained version; and this demonstrates the benefits of our dataset. The state-of-the-art few-shot detectors~\cite{kang2019few,fan2020few} perform relatively poor, even when they are trained on our dataset.
With these results, we are the first to show the empirical evidence for the inferiority of the detection-then-counting approach compared to the density estimation approach (GMN, MAML, FamNet) for generic object counting. However, this is not new for the crowd counting research community, where the density estimation approach dominates the recent literature~\cite{zhang2016single}, thanks to its robustness to occlusion and the freedom of not having to commit to binarized decisions at an early stage. Among the competing approaches, MAML is the best method of all. This is perhaps because MAML is a meta learning method that leverages the advantages of having the FamNet architecture as its core component. The MAML way of training this network leads to a better model than GMN, but it is still inferior to the proposed FamNet together with the proposed training and adaptation algorithms. In terms of training time per epoch, FamNet is around three times faster than MAML, because it does not require any higher order gradient computation like MAML. 

%%%%%%%%%%%%%%%%%%%%%%%%%%%%%%%%%%%%%%%%%%%%%%%%%%%%%%%%%%%%%%%%%%%%%%%%%%%
%%%%%%%%%%%%%%%%%%%%%%%%%%%%%%%%%%%%%%%%%%%%%%%%%%%%%%%%%%%%%%%%%%%%%%%%%%%%%

\setlength{\tabcolsep}{3pt}
\begin{table}[!t]
%\vskip 0.1in
\centering
\begin{tabular}{lcccc}
\toprule
          &  \multicolumn{2}{c}{ Val-COCO Set} &  \multicolumn{2}{c}{ Test-COCO Set} \\
         \cmidrule(lr){2-3} \cmidrule(lr){4-5} 
    Method &  MAE          & RMSE    & MAE          & RMSE         \\

\midrule 
Faster R-CNN & 52.79 & 172.46 & 36.20 & 79.59 \\
RetinaNet & 63.57 & 174.36 & 52.67 & 85.86 \\
Mask R-CNN & 52.51 & 172.21 & 35.56 & 80.00 \\
%ExamNet (No Adaptation) & 49.03 & 133.32 & 29.29 & 57.80 \\
FamNet (Proposed)  & \textbf{39.82} & \textbf{108.13} & \textbf{22.76} & \textbf{45.92}  \\
%FamNet (Proposed)  & \textbf{41.49} & \textbf{126.08} & \textbf{23.02} & \textbf{51.20}  \\
%\color{red}{
%FamNet (Proposed)}  & \textbf{47.62} & \textbf{119.37} & \textbf{24.49} & %\textbf{50.71}  \\

\bottomrule 
\end{tabular}
\vskip -0.1in
\caption{{\bf Comparing FamNet with pre-trained object detectors}, on counting objects from categories where there are pre-trained object detectors. 
\label{tab:detectors}}
\end{table}

\subsection{Comparison with Object Detectors}
One approach for counting is to use a detector to detect objects and then count. This approach only works for certain categories of objects, where there are detectors for those categories. In general, it requires thousands of examples to train an object detector, so this is not a practical method for general visual counting. Nevertheless, we evaluate the performance of FamNet on a subset of categories from the validation and test sets that have pre-trained object detectors on the COCO dataset. We refer to these subsets as Val-COCO and Test-COCO, which comprise of 277 and 282 images respectively. Specifically, we compare FamNet with FasterRCNN~\cite{Ren-etal-NIPS15}, MaskRCNN~\cite{He-etal-ICCV17}, and RetinaNet~\cite{lin2017focal}. All of these pretrained detectors are available in the Detectron2 library~\cite{wu2019detectron2}. \Tref{tab:detectors} shows the comparison results. As can be seen, FamNet outperforms the pre-trained detectors, even on object categories where the detectors have been trained with thousands of annotated examples from the COCO dataset. 
%\mhoai{Why are the results much worse for Val?}
%.Val-COCO consists bird, skateboard, book, horse and chair classes and Test-COCO consists of apple, sheep, elephant and skis classes.
%%%%%%%%%%%%%%%%%%%%%%%%%%%%%%%%%%%%%%%%%%%%%%%%%%%%%%%%%%%%%%%%%%%%%%%%%%%
%%%%%%%%%%%%%%%%%%%%%%%%%%%%%%%%%%%%%%%%%%%%%%%%%%%%%%%%%%%%%%%%%%%%%%%%%%%%%
\setlength{\tabcolsep}{10pt}
\begin{table}[!t]
%\vskip 0.1in
\centering
\begin{tabular}{crr}
\toprule
Number of Exemplars & MAE & RMSE \\
\midrule 
1 & 26.55 & 77.01 \\
2 & 24.09 & 72.37 \\
3 & 23.75 & 69.07 \\
%1 &  &  \\
%2 &  &  \\
%3 & 22.76 & 75.81 \\
\bottomrule 
\end{tabular}
\vskip -0.1in
\caption{{Performance of FamNet on the validation data as the number of exemplars increases}. FamNet can provide a reasonable count estimate even with a single exemplar, and the estimate becomes more accurate with more exemplars.  
\label{tab:NoExemplars}}
\end{table}

\iffalse
\setlength{\tabcolsep}{10pt}
\begin{table}[!t]
%\vskip 0.1in
\centering
\begin{tabular}{crr}
\toprule
Number of Exemplars & MAE & RMSE \\
\midrule 
1 & 28.49 & 86.21\\
2 & 27.22 & 80.40 \\
3 & 26.80 & 73.83 \\
\bottomrule 
\end{tabular}
\vskip -0.1in
\caption{{\vr{entire table}Performance of FamNet on the validation data as the number of exemplars increases}. FamNet can provide a reasonable count estimate even with a single exemplar, and the estimate becomes more accurate with more exemplars.  
\label{tab:NoExemplars}}
\end{table}

\setlength{\tabcolsep}{3pt}
\begin{table}[!thb]
\centering
\begin{tabular}{lcccc}
\toprule
Components & \multicolumn{4}{c}{Combinations}   \\
\midrule
Multi-scale image feature  & \xmark  & \checkmark &\checkmark &\checkmark \\
Multi-scale exemplar feature & \xmark & \xmark &\checkmark & \checkmark \\
Test time adaptation & \xmark& \xmark& \xmark& \checkmark  \\
\midrule 
%MAE & 31.02 & 30.70 & 27.90 & 26.80 \\
%RMSE & 101.6 &  95.93 & 79.78 & 73.83 \\
MAE & 31.02 & 30.70 & 27.90 & 26.80 \\
RMSE & 101.6 &  95.93 & 79.78 & 73.83 \\
%\mhoai{What is Exemplar feature augmentation????}
\bottomrule
\end{tabular}
\vskip -0.1in
\caption{\vr{entire table}{\bf Analyzing the components of FamNet}. Each of the components of FamNet adds to the performance. \label{tab:ablation}}
\end{table}
\fi

\setlength{\tabcolsep}{3pt}
\begin{table}[!thb]
\centering
\begin{tabular}{lcccc}
\toprule
Components & \multicolumn{4}{c}{Combinations}   \\
\midrule
Multi-scale image feature & \xmark & \checkmark &\checkmark & \checkmark \\
Multi-scale exemplar feature  & \xmark  & \xmark &\checkmark &\checkmark \\
Test time adaptation & \xmark& \xmark& \xmark& \checkmark  \\
\midrule 
%MAE & 31.02 & 30.70 & 27.90 & 26.80 \\
%RMSE & 101.6 &  95.93 & 79.78 & 73.83 \\
% with max size exemplar
%MAE & 29.71 & 27.39 & 25.48 & 24.97 \\
%RMSE & 100.63 &  89.54 & 79.39 & 79.34 \\
MAE & 32.70 & 27.80 & 24.32 & 23.75 \\
RMSE & 104.31 &  93.53 & 70.94 & 69.07 \\
%MAE & 32.70 & 27.80 & 23.50 & 22.76 \\
%RMSE & 104.31 &  93.53 & 79.91 & 75.81 \\
%\mhoai{What is Exemplar feature augmentation????}
\bottomrule
\end{tabular}
\vskip -0.1in
\caption{{\bf Analyzing the components of FamNet}. Each of the components of FamNet adds to the performance. \label{tab:ablation}}
\end{table}

\subsection{Ablation Studies}
We perform ablation studies on the validation set of FSC-147 to analyze: (1) how the counting performance changes as the number of exemplars increases, and (2) the benefits of different components of FamNet.

%\myheading{Counting error as the number of exemplars increases.}
In \Tref{tab:NoExemplars}, we analyze the performance of FamNet as the number of exemplars is varied between one to three during the testing of FamNet. We see that FamNet can work even with one exemplar, and it outperforms all the competing methods presented in \Tref{tab:baseline} with just 2 exemplars. Not surprisingly, the performance of FamNet improves as the number of exemplars is increased. This suggests that an user of our system can obtain a reasonable count even with a single exemplar, and they can obtain a more accurate count by providing more exemplars. 

%can provide more exemplars if they are interested in having a more accurate count. 

In \Tref{tab:ablation}, we analyze the importance of the key components of FamNet: multi-scale image feature map, the multi-scale exemplar features, and test time adaptation. We train models without few/all of these components on the training set of FSC-147, and report the validation performance. We notice that all of the components of FamNet are important, and adding each of the component leads to improved results.

%%%%%%%%%%%%%%%%%%%%%%%%%%%%%%%%%%%%%%%%%%%%%%%%%%%%%%%%%%%%%%%%%%%%%%%%%%%
%%%%%%%%%%%%%%%%%%%%%%%%%%%%%%%%%%%%%%%%%%%%%%%%%%%%%%%%%%%%%%%%%%%%%%%%%%%%%
\subsection{Counting category-specific objects}

FamNet is specifically designed to be general, being able to count generic objects with only a few exemplars. As such, it might not be fair to demand it to work extremely well for a specific category, such as counting cars. Cars are popular objects that appear in many datasets and this category is the explicit or implicit target for tuning for many networks, so it would not be surprising if our method does not perform as well as other customized solutions. Having said that, we still investigate the suitability of using FamNet to count cars from the CARPK dataset~\cite{hsieh2017drone}, which consists of overhead images of parking lots taken by downward facing drone cameras. The training and test set consists of 989 and 459 images respectively. There are around 90,000 instances of cars in the dataset. 

We experiment with two variants of FamNet: a pre-trained model and a model trained on CARPK dataset. The pre-trained FamNet model is called FamNet--, which is trained on FSC-147, without using the data from CARPK or the car category from FSC-147. The FamNet model trained with training data from CARPK is called FamNet+, and it is trained as follows. 
We randomly sample a set of 12 exemplars from the training set, and use these as the exemplars for all of the training and test images. We train FamNet$+$ on the CARPK training set. \Tref{tab:carspk} displays the results of several methods on this CARPK dataset.
FamNet+ outperforms all methods except GMN~\cite{lu2018class}. GMN, unlike all the other approaches, uses extra training data from the ILSVRC video dataset which consists of video sequences of cars. Perhaps this may be why GMN works particularly well on CARPK.

%Since one of the training categories in FSC-147 dataset is cars, we remove all images of cars category from the training set for this experiment, and train FamNet from scratch. We notice that the pre-trained FamNet outperforms several of the previous methods which were trained on the entire CARSPK training set comprising of 989 images. For the FamNet$+$ variant, 

\setlength{\tabcolsep}{3pt}
\begin{table}[!tb]
\vskip .1in
\centering
\begin{tabular}{lrrrr}
\toprule
Method & MAE & RMSE \\
\midrule 
YOLO~\cite{Redmon-et-al-CVPR16,hsieh2017drone} & 48.89 & 57.55 \\
Faster RCNN~\cite{ren2015faster,hsieh2017drone} & 47.45 & 57.39 \\
One-look Regression~\cite{mundhenk2016large,hsieh2017drone} & 59.46 & 66.84 \\
Faster RCNN~\cite{ren2015faster,hsieh2017drone}(RPN-small) & 24.32 & 37.62 \\
Spatially Regularized RPN~\cite{hsieh2017drone} & 23.80 &36.79 \\
GMN \cite{lu2018class} & 7.48 & 9.90  \\
%ExamNet (No Adaptation)   & 36.22 &  52.19 \\
FamNet-- (pre-trained)    & 28.84 &  44.47 \\
FamNet+ (trained with CARPK data) & 18.19 & 33.66 \\
%FamNet+ (trained with CARPK data) & 17.47 & 34.92 \\
\bottomrule 
\end{tabular}
\vskip -0.1in
\caption{{\bf Counting car performance on the CARPK dataset}. FamNet-- is a FamNet model, that is trained without any CARPK images nor images from the car category of FSC-147. Other methods use the entire CARPK train set. Pre-trained FamNet-- outperforms three of of the previous approaches. FamNet+, yields even better performance. 
\label{tab:carspk}}
\end{table}

\newcommand\qualextwo{0.23\textwidth}
\newcommand\qualexthree{0.15\textwidth}
\newcommand\heightqualex{2.2cm}
\begin{figure}[h]  
\centering
\makebox[\qualextwo]{Image}
\makebox[\qualextwo]{Prediction}

    \includegraphics[width=\qualextwo,height=\heightqualex]{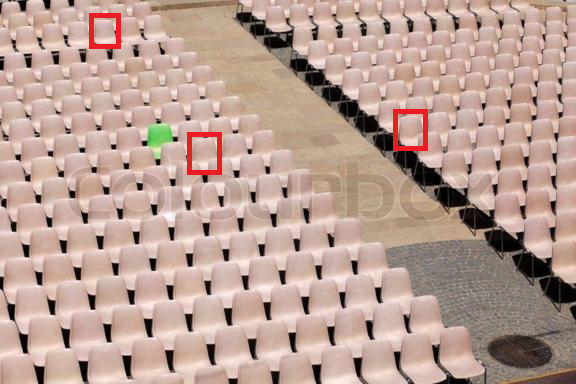}
    \includegraphics[width=\qualextwo,height=\heightqualex]{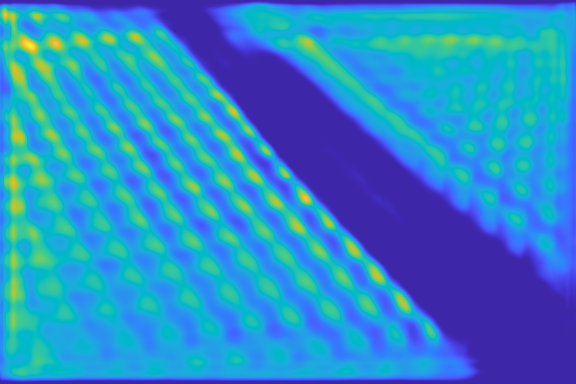} \\    
    % \makebox[\qualextwo]{(a) Input image}
\makebox[\qualextwo]{GT Count: 263}
\makebox[\qualextwo]{Pred Count: 280} \\ \vspace{2ex}

    \includegraphics[width=\qualextwo,height=\heightqualex]{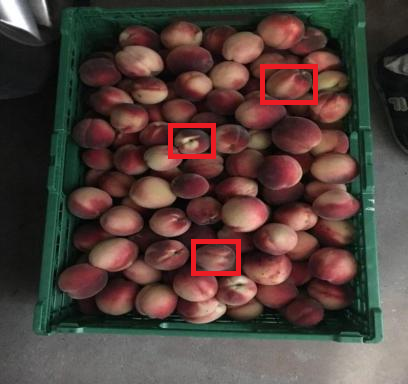}
    \includegraphics[width=\qualextwo,height=\heightqualex]{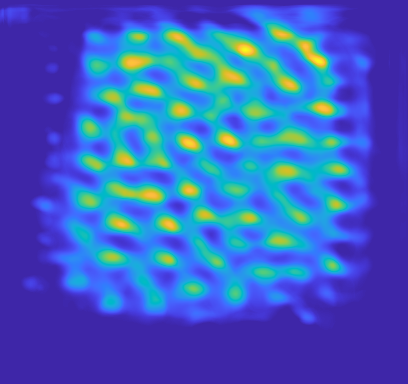} \\    
    % \makebox[\qualextwo]{(a) Input image}
\makebox[\qualextwo]{GT Count: 77}
\makebox[\qualextwo]{Pred Count: 77} \\ \vspace{2ex}

    \includegraphics[width=\qualextwo,height=\heightqualex]{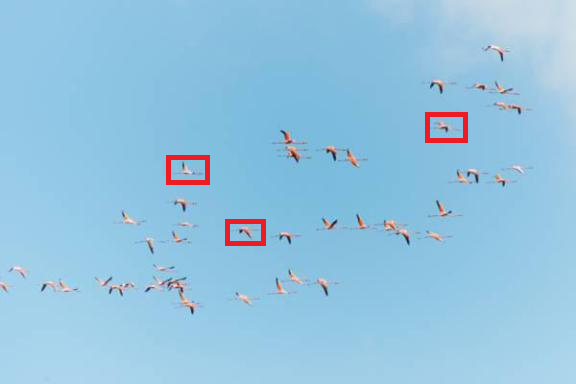}
    \includegraphics[width=\qualextwo,height=\heightqualex]{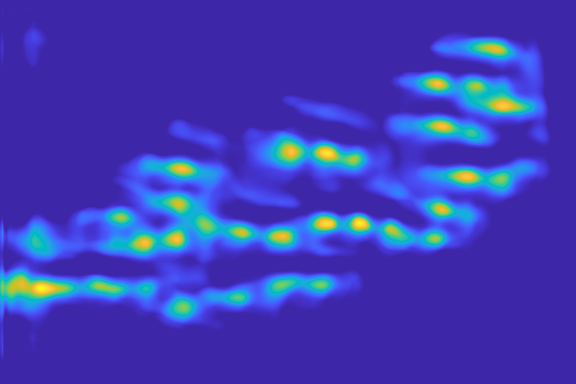} \\    
    % \makebox[\qualextwo]{(a) Input image}
\makebox[\qualextwo]{GT Count: 47}
\makebox[\qualextwo]{Pred Count: 46} \\ \vspace{2ex}

    \includegraphics[width=\qualextwo,height=\heightqualex]{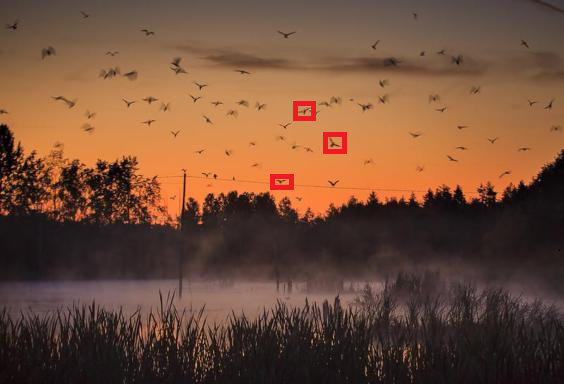}
    \includegraphics[width=\qualextwo,height=\heightqualex]{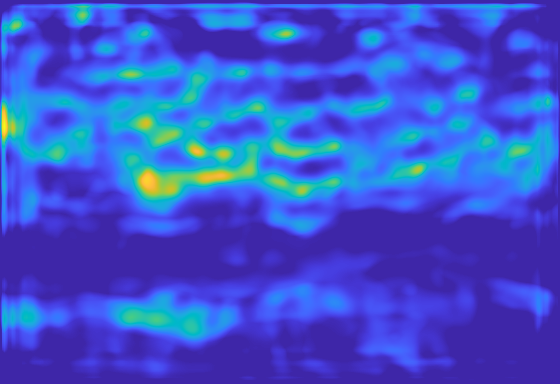} \\    
    % \makebox[\qualextwo]{(a) Input image}
\makebox[\qualextwo]{GT Count: 77}
\makebox[\qualextwo]{Pred Count: 192} \\ \vspace{2ex}
\iffalse
    \includegraphics[width=\qualexthree,height=\heightqualex]{Images/Quali/2243_image.png}
    \includegraphics[width=\qualexthree,height=\heightqualex]{Images/Quali/2243_pred.png}
        \includegraphics[width=\qualexthree,height=\heightqualex]{Images/Quali/2243_pred_adapt.png}\\    
    % \makebox[\qualextwo]{(a) Input image}
\makebox[\qualexthree]{GT Count: 544}
\makebox[\qualexthree]{Count: 147} 
\makebox[\qualexthree]{Count: 177}\\ \vspace{2ex} \fi
  \vskip -0.1in
  \caption{{\bf Predicted density maps and counts of FamNet}. }
  \label{fig:Qualitative}
\end{figure}

\begin{figure}[h]  
\centering
\makebox[\qualexthree]{Image}
\makebox[\qualexthree]{Pre Adapt.}
\makebox[\qualexthree]{Post Adapt.}
    \includegraphics[width=\qualexthree,height=1.8cm]{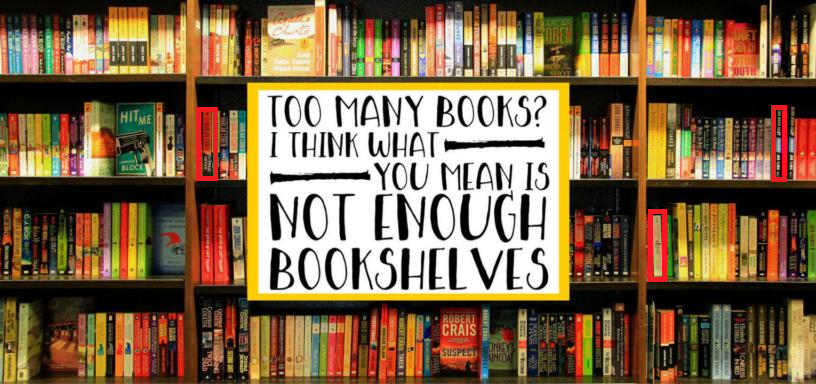}
    \includegraphics[width=\qualexthree,height=1.8cm]{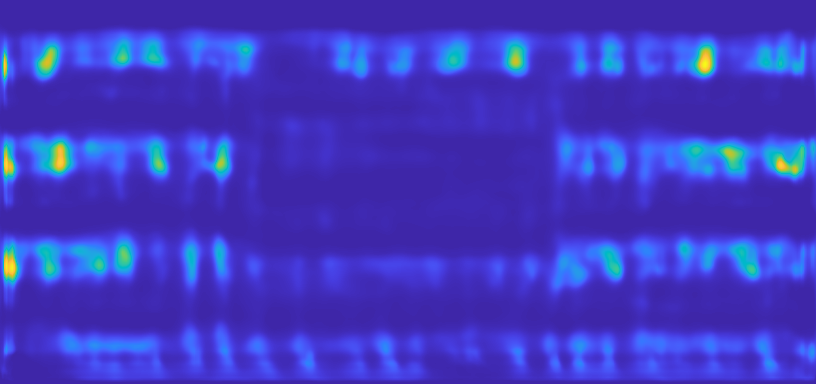}
        \includegraphics[width=\qualexthree,height=1.8cm]{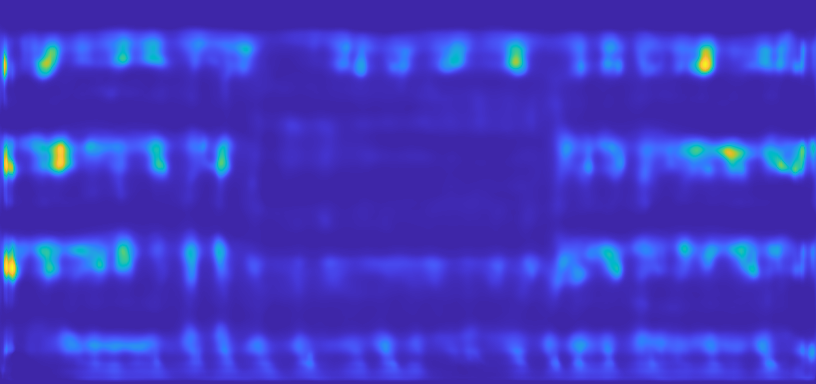}\\    
    % \makebox[\qualextwo]{(a) Input image}
\makebox[\qualexthree]{GT Count: 240}
\makebox[\qualexthree]{Count: 356} 
\makebox[\qualexthree]{Count: 286}\\ \vspace{2ex}
  \vskip -0.1in
  \caption{{\bf Test time adaptation}. Shown are the initial density map (Pre Adapt) and final density map after adaptation (Post Adapt). In case of over counting, adaptation decreases the density values at dense locations. }
  \label{fig:Qualitative2}
\end{figure}

\subsection{Qualitative Results}
\Fref{fig:Qualitative} shows few images and FamNet predictions. The first three are success cases,and the last is a failure case. For the fourth image, FamNet confuses portions of the background as being the foreground, because of similarity in appearance between the background and the object of interest. \Fref{fig:Qualitative2} shows a test case where test time adaptation improves on the initial count by decreasing the density values in the dense regions.

\section{Conclusions}
In this paper, we posed counting as a few-shot regression task. Given the non-existence of a suitable dataset for the few-shot counting task, we collected a visual counting dataset with relatively large number of object categories and instances. We also presented a novel approach for density prediction suitable for the few-shot visual counting task. We compared our approach with several state-of-art detectors and few shot counting approaches, and showed that our approach outperforms all of these approaches.

\myheading{Acknowledgements:} This project is partially supported by MedPod, the SUNY2020 Infrastructure Transportation Security Center, and the NSF I/UCRC Center for Visual and Decision Informatics at Stony Brook.

{\small
\bibliographystyle{ieee_fullname}
\setlength{\bibsep}{0pt}

\bibliography{longstrings,egbib,pubs}
}

\end{document}